\documentclass[runningheads]{llncs}

\usepackage{graphicx}
\usepackage{subfig}
\usepackage{amssymb}
\usepackage{amsmath}
\usepackage{multirow,tabularx}
\usepackage{hhline}
\usepackage{enumitem}
\usepackage{subfig}
\usepackage{url}
\usepackage{amssymb}
\usepackage{url}
\usepackage{ragged2e} 
\usepackage{adjustbox}
\usepackage{setspace}
\usepackage{rotating}
\usepackage{pdfpages}
\usepackage{caption}

\usepackage{epstopdf}

\begin{document}
\title{Surgical Gesture Recognition with Optical Flow only}
%
%
\author{Duygu Sarikaya\inst{1}\orcidID{0000-0002-2083-4999} \and
Pierre Jannin\inst{1} }
\authorrunning{F. Author et al.}
%
\institute{University of Rennes 1, Inserm, LTSI, Rennes, France}
\maketitle              
\begin{abstract}
In this paper, we address the open research problem of surgical gesture recognition using motion cues from video data only. We adapt Optical flow ConvNets initially proposed by Simonyan \textit{et al.}. While Simonyan uses both RGB frames and dense optical flow, we use only dense optical flow representations as input to emphasize the role of motion in surgical gesture recognition, and present it as a robust alternative to kinematic data. We also overcome one of the limitations of Optical flow ConvNets by initializing our model with cross modality pre-training. A large number of promising studies that address surgical gesture recognition highly rely on kinematic data which requires additional recording devices. To our knowledge, this is the first paper that addresses surgical gesture recognition using dense optical flow information only. We achieve competitive results on JIGSAWS dataset, moreover, our model achieves more robust results with less standard deviation, which suggests optical flow information can be used as an alternative to kinematic data for the recognition of surgical gestures. 

\end{abstract}
\section{Introduction}

Surgical activity recognition is an active research area. The potential of autonomous or human-robot collaborative surgeries, automated real-time feedback, guidance and navigation during surgical tasks is exciting for the community. There are studies that address this open research problem. The methods range from using SVM classification with Linear Discriminant Analysis (LDA) \cite{lin2005,lin2006}, Gaussian mixture models (GMMs) \cite{leong},  variations of Hidden Markov Models (HMM) that represents probability distributions over sequences of observations, \cite{yang,varadarajan} and more recently, to convolutional neural networks and recurrent neural networks \cite{m2cai_smooth,colin,m2cai_lstm}. 

DiPietro \textit{et al.} \cite{colin} propose using Recurrent Neural Networks (RNN) trained on kinematic data for surgical gesture classification on JIGSAWS dataset \cite{jigsaws}. Some other recent studies focus on classifying surgery phases; a higher level of surgical activity that includes a sequence of different tasks. A recent work by Cad\'ene \textit{et al}. \cite{m2cai_smooth} uses deep residual networks to extract visual features of the video frames and then applies temporal smoothing with averaging. The authors of this work finally model the transitions between the surgery phase steps with an HMM. Twinanda \textit{et al.} \cite{m2cai_lstm} offer a study on classification of surgery phases by first extracting visual features of video frames via a CNN and then passing them to an SVM to compute confidences of a frame belonging to a surgery phase. Sarikaya \textit{et al.} \cite{sarikaya} proposes a multi-modal and multi-task CNN-LSTM architecture for simultaneous gesture and surgical task classification, they argue that  surgical tasks are better modeled by the visual features that are determined by the objects in the scene, while gestures are better modeled with motion cues, however their performance remains on the lower end. Similary, we argue that surgical gestures are best modeled with motion cues as these cues can be charaterized by signature motions of surgeons. 

Going through the more recent citations on JIGSAWS benchmark, we noticed that a large number of promising studies highly rely on kinematic data. Kinematic data requires additional recording devices, while a computer vision approach doesn't.  We propose a computer vision only approach by using dense optical flow as input as an alternative to kinematic data. We demonstrate that using optical flow information only, we achieve competitive results. While the setting and objects differ across different tasks, the surgeon's motion and the transitions between these motions remain generic.  For example, the common gesture \textit{Positioning the needle} \ref{table:gestures} might take place in different tasks of \textit{Suturing} and \textit{Needle Passing} \cite{jigsaws}. While the settings and objects differ for these tasks, the gesture \textit{Positioning the needle} motions are identical and can be identified by the temporal dynamics. In this way, gestures can be defined with motion cues which are independent of the setting and the objects. Similarly, to differentiate between gestures in the same setting,  we can use motion as a reliable identifier while the advantage of visual cues to recognize gestures is not so evident in such setting. In this paper, we adapt Optical flow ConvNets initially proposed by Simonyan \textit{et al.} \cite{zisserman}. While Simonyan \textit{et al.} uses both RGB frames and dense optical flow, we use only dense optical flow representations as input to emphasize the role of motion in surgical gesture recognition and present it as a robust alternative to kinematics and RGB frames. We also overcome one of the limitations of Optical flow ConvNets;  Simonyan \textit{et al.}  initializes the weights of their spatial network on RGB frames with a pretrained model on Imagenet, however they do not initialize the weights of their Optical flow ConvNets for the lack of an alternative. We overcome this limitation by initializing our Optical flow ConvNets with the method called cross modality pre-training proposed by Wang \textit{et al.} \cite{wang}. Using the pretrained Imagenet weights on RGB, we first average the weight value across the RGB channels and replicate this average by the channel number of motion stream input. This helps our model to better converge and avoid overfitting, which is a significant gain since we work with JIGSAWS \cite{jigsaws} which is a small dataset (and as far as we are concerned, the only public dataset that provides gesture annotation, which are low-level atomic surgical activities). Using a simple model as suggested we were able to get competitive results on JIGSAWS gesture classification task. We experiment with our model using JIGSAWS's Leave-one-supertrial-out (LOSO) cross-validation scheme, and compare our results to the benchmark \cite{jigsaws_benchmark}.

\begin{figure}
\centering
\subfloat{\includegraphics[width = 5in]{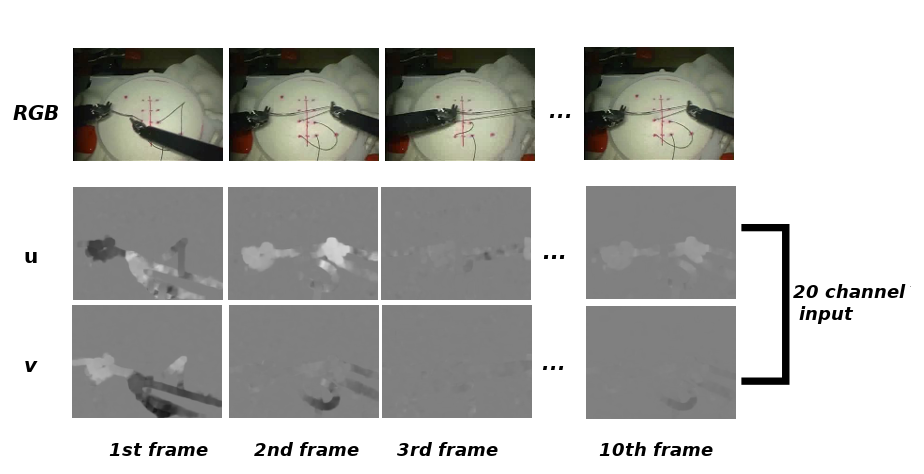}}
\caption{ Our input to the BN-ResNet101 architecture is a 2-channel array of dense optical flow vectors corresponding to video frames. $(u,v)$ represents the horizontal and vertical components of the vector field \cite{zisserman}. So for a multi-frame length of $10$, our input is a total of 20 channels ($L=10*2-channels(u,v)$). Our BN-ResNet101 model then classifies videos with surgical gesture labels. }
 \label{fig:method}
\end{figure}

\section{Methods} \label{section_methods}

\subsection{Optical flow ConvNets}

Our choice of using dense optical flow for gesture recognition is supported by the findings of Karpathy  \cite{karpathy} \textit{et al.} that shows a network operating on individual RGB video frames performs similarly to the networks whose input is a stack of RGB frames, and thus learnt spatio-temporal features from these RGB frame models do not capture the motion well. Similar to the architecture proposed by Simonyan \textit{et al.} \cite{zisserman} (Optical flow ConvNets), the input to our model is formed by stacking dense optical flow displacement fields between several consecutive frames (we chose the length of consecutive frames as $L=10$ in accordance with the proposed model \cite{zisserman} ). The dense optical flow is computed using Farneback method \cite{farneback} by solving for a displacement field at multiple image scales. We chose to work with Farneback as its computation is fast.  We get a 2-channel array with optical flow vectors, $(u,v)$ that represents the horizontal and vertical components of the vector field correspondingly \cite{zisserman}, which results in a total of 20 channels for each input (\textit{L=$10*2$-channels$(u,v)$}) (Figure \ref{fig:method}). We find their magnitude and direction, and save these two separate images as grayscale image representations rescaled to a [0, 255] range and compressed using JPEG. We adapt a BN-ResNet101, a ResNet with Batch Normalization (BN) where BN is used for normalizing the value distribution before going into the next layer \cite{ResNet,BN}. Since we work with a small dataset, we also add a dropout layer after pooling of ResNet to overcome overfitting. Normally, BN addresses this problem, however while working with smaller datasets, using BN only could lead to overfitting. We also adapt additional normalization techniques such as scaling the dense optical flow grayscale representations to $[0.1]$, and data augmentation techniques such as taking five random crops for each frame. Also, since we take \textit{L=$10$ consequent frames} of chunks for each video, it could be considered as further data augmentation, a means of random cropping for videos similar to random cropping of images \cite{Donahue}. As mentioned earlier, we initialize our Optical flow ConvNets with the method called cross modality pre-training proposed by Wang \textit{et al.} \cite{wang}. Using the pretrained Imagenet weights on RGB, we first average the weight value across the RGB channels and replicate this average by the channel number of motion stream input, which in our case is $20$. This helps us greatly to overcome overfitting and addresses the limitations of Optical flow ConvNets as initially suggested by Simonyan \textit{et al.}. 

\section{Experiments}
\subsection{Dataset} 
\begin{figure}
\centering
\subfloat{\includegraphics[width = 1.66in]{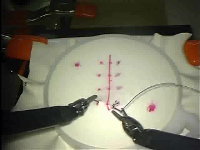}}
\subfloat{\includegraphics[width = 1.66in]{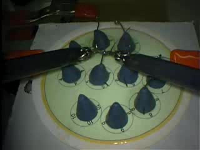}}
\subfloat{\includegraphics[width = 1.66in]{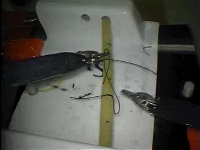}}
\caption{ The JHU-ISI Gesture and Skill Assessment Working Set (JIGSAWS) \cite{jigsaws} provides a public benchmark surgical activity dataset. $8$ surgeons with varying expertise performs $3$ surgical tasks: (from left to right) \textit{Suturing}, \textit{Needle Passing} and \textit{Knot Tying} on the daVinci Surgical System (dVSS $^{\tiny{\textregistered}}$). }
 \label{fig:JHU}
\end{figure}

The JHU-ISI Gesture and Skill Assessment Working Set (JIGSAWS) \cite{jigsaws} provides a public benchmark surgical activity dataset. In this video dataset, $8$ surgeons with varying expertise perform $3$ surgical tasks on the daVinci Surgical System (dVSS $^{\tiny{\textregistered}}$)  (Figure \ref{fig:JHU}). The dataset includes video data captured from endoscopic cameras of the dVSS at $30$ Hz and at a resolution of $640x480$. The videos are recorded during the performance of the tasks: \textit{Suturing}, \textit{Needle Passing} and \textit{Knot Tying}, and the dataset provides $15$ low-level gesture labels, which are the smallest action units where the movement is intentional and is carried out towards achieving a specific goal. The gestures form a common vocabulary for small action segments that reoccur in different tasks. A list of the gestures is given in the Table \ref{table:gestures}.

\begin{figure}
\centering
\subfloat{\includegraphics[width = 1in]{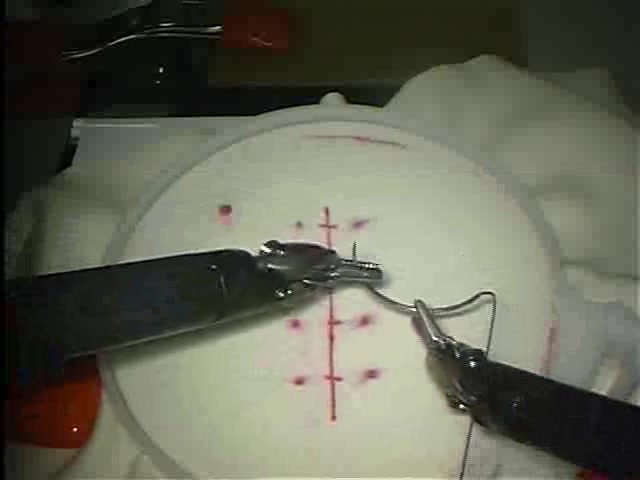}\includegraphics[width = 1in]{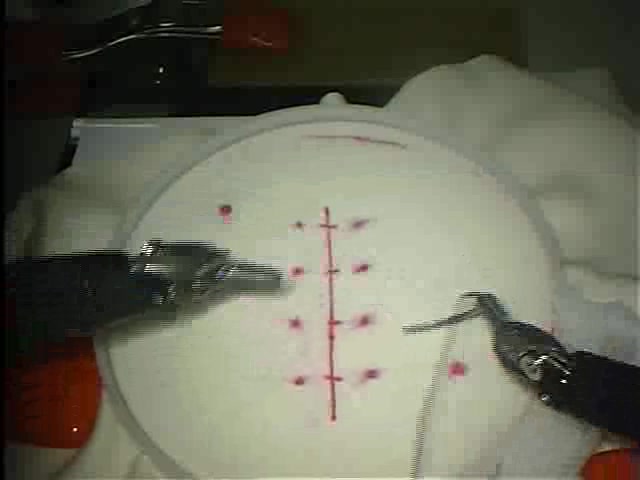}\includegraphics[width = 1in]{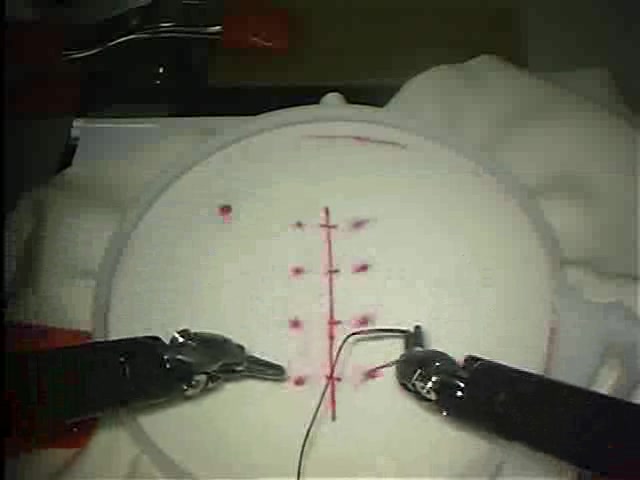}\includegraphics[width = 1in]{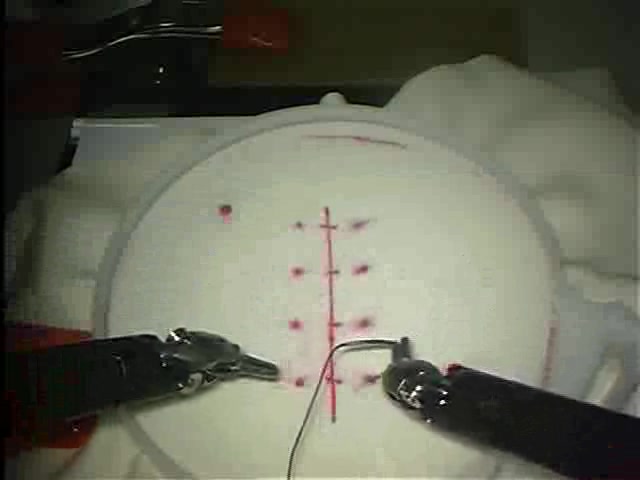}\includegraphics[width = 1in]{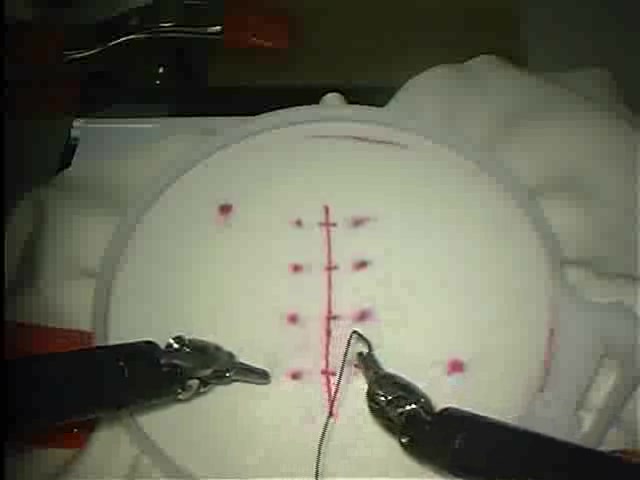}} \\
\includegraphics[width = 1in]{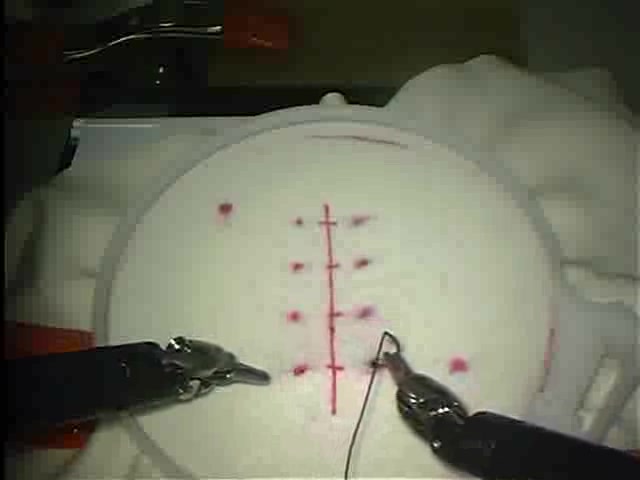}\includegraphics[width = 1in]{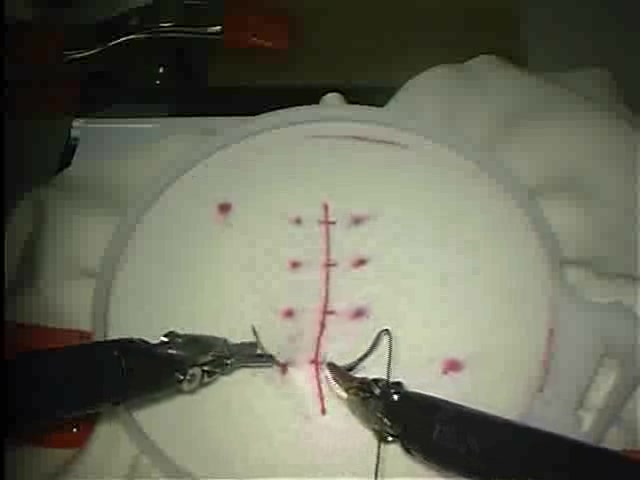}\includegraphics[width = 1in]{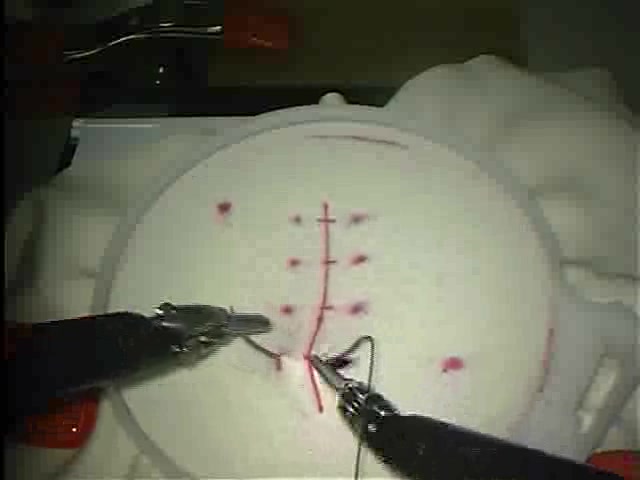}\includegraphics[width = 1in]{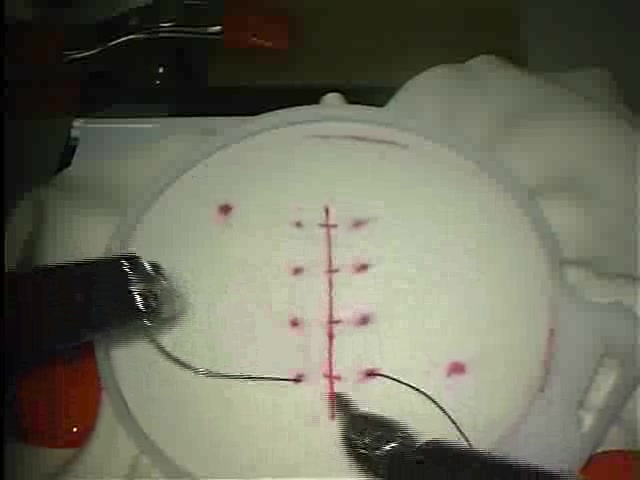}\includegraphics[width = 1in]{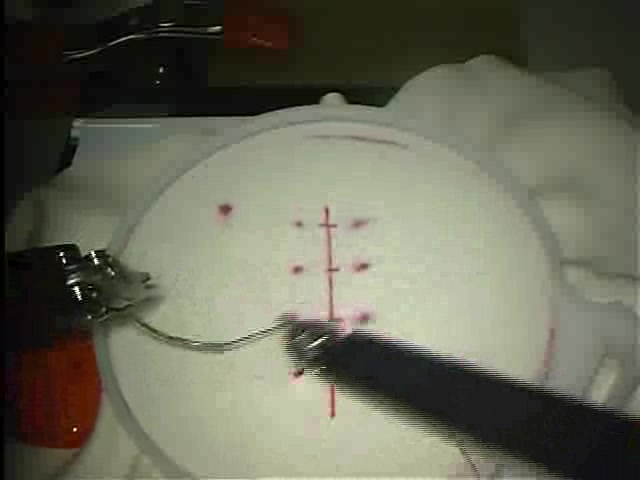} \\
\caption{ A temporal sequence of surgical activities during a \textit{Suturing} task are shown (from left to right, top to bottom). The surgeon does a suture on the tissue following the guide landmarks. }
 \label{fig:jigsaws_suturing}
\end{figure}

\begin{table}
\centering
\caption{Gesture vocabulary}\label{table:gestures}
\begin{tabular}{|l|l|l|}
\hline
Heading level &  Example \\
\hline
G1 &   Reaching for needle with right hand   \\
G2 & Positioning needle   \\
G3 & Pushing needle through tissue \\
G4 &  Transferring needle from left to right\\
G5 &  Moving to center with needle in grip    \\
G6 &   Pulling suture with left hand        \\
G7 & Pulling suture with right hand   \\
G8 & Orienting needle    \\
G9 &  Using right hand to help tighten suture \\
G10 &  Loosening more suture    \\
G11 &   Dropping suture at end and moving to end points   \\
G12 & Positioning needle   \\
G13 & Making C loop around right hand     \\
G14 &  Reaching for suture with right hand \\
G15 &  Pulling suture with both hands    \\
\hline
\end{tabular}
\end{table}

\subsection{Preprocessing Data}
We first clip the task videos into gesture clips using the start and end frame information for each gesture. Then, we extract each RGB frame (resized to $320 x 240$) at $30Hz$. For clips with a duration less than $30$ frames (less than $1$ second), we extract the frames at $40Hz$ instead, this way we partially balance the dataset. We excluded the few clips resulting in having less than $15$ frames, we have also noticed that they might be wrongly annotated as a gesture loop is not completed in the timeframes given. We rescale our input data to $256x256$ and take a random of five crops with the size $224x224$.

After extracting the frames, we have computed the dense optical flow of each consequent frame using the method described by Farneback \cite{farneback}. Although there are more recent and precise methods, we chose to work with Farneback method as it is fast.  We get a $2-channel$ array with optical flow vectors, $(u,v)$ that represents the horizontal and vertical components of the vector field \cite{zisserman}.  We find their magnitude and direction, and save these two separate images as grayscale image representations as mentioned in \textit{Methods}. 

\subsection{Experimental Setup}
JIGSAWS come with an experimental setup that can be used to evaluate automatic surgical gesture classification methods. Leave-one-supertrial-out (LOSO) is one of the cross-validation schemes included in this setup. According to this scheme, Supertrial $i$ is defined as the set consisting of the $i$-th trial from all subjects for a given surgical task. In the LOSO setup for cross-validation, five folds with each fold comprising of data from one of the five supertrials are provided. The LOSO setup can be used to evaluate the robustness of a method by leaving out the $i$-th repetition for all subjects. Although this setup has \textit{Train} and \textit{Test} splits for each fold, it does not provide a separate \textit{Validation} split. To ensure robustness of our model, and to avoid overfitting, we split the \textit{Train} list once again to leave a random trial out, and we use that trial as the \textit{Validation} set. This way, we make sure that no videos belonging to the same trial appears in both \textit{Train} and \textit{Validation}.

We carried out our experiments with a \textit{TITAN X (Pascal architecture) GPU} and an  \textit{Intel Xeon (R) CPU E5 3.50 GHz x8} with a \textit{31.2 GiB} memory. We use python for implementation in general and pytorch in particular for deep learning. We use a learning rate of $1e-3$ with a Stochastic gradient descent (SGD) optimizer, and we decrease this learning rate in steps ( \textit{stepsize=$10$, gamma=$0.25$}). Our main focus was to overcome overfitting and ensure a network that can generalize, in addition to this we did experiment in a grid search manner to further optimize our model, however for the latter, there is room for improvement and a more broad grid-search could be done. For testing, we uniformly sample $20$ frames in each video and the video level prediction is the voting result (averaging) of all $20$ frame level predictions. Our training takes only about $20 seconds$ per mini-batch, which we have set at $30$, and an epoch for testing takes about $25-30$ seconds for $80-100$ $Suturing$ videos which are the longest in the dataset, and about $10-15$ seconds for $Knot Tying$ videos which are shorter.  We train our models for $300$ epochs, however they converge around $80-150$th epoch.

The methods we used are easy to reimplement as authors of mentioned methods Optical flow ConvNets, cross modality pre-training, ResNet, BN \cite{zisserman,wang,ResNet,BN} share their code publicly. Dense optical flow computation \cite{farneback} can be found in OpenCV's library. In the future, we may release additional scripts, experimentation splits and information on the 3rd party software used to make it easier to reproduce the experiments. 

\subsection{Results and Conclusion}

Our experimentation results on JIGSAWS dataset are shown in Table \ref{tab1}. Our optical flow ConvNets model using dense optical flow information only performs well and with small standard deviation. In addition, our training takes only about $20 seconds$ per mini-batch, which we have set at $30$, and an epoch for testing takes about $25-30$ seconds for $80-100$ \textit{Suturing} videos which are the longest in the dataset, and about $10-15$ seconds for \textit{Knot Tying} videos which are shorter. Our model can be extended for activity segmentation, and our competitive results suggest that optical flow information could be used as an alternative to kinematic data. 

\begin{table}
\caption{Classification Techniques Validated on the
JIGSAWS, for LOSO Cross Validation}\label{tab1}
\begin{tabular}{|l|l|l|l|l|}
\hline
Method (Data Type)&  Evaluation  & Suturing & Needle Passing & Knot Tying\\
\hline
Our model (vid/opt flow) & Accuracy $\pm$ std & 91.07  $\pm$ 0.67 & 74.25 $\pm$ 3.66 & 87.78 $\pm$ 3.44 \\
\hline
\end{tabular}
\end{table}

\textbf{Acknowledgements} This work was supported by French state funds managed within the
Investissements d’Avenir program by BPI France (project CONDOR).

%
%
%

\end{document}